\pdfoutput=1

\documentclass[11pt]{article}

\usepackage[preprint]{acl}

\usepackage{times}
\usepackage{latexsym}

\usepackage[T1]{fontenc}

\usepackage[utf8]{inputenc}

\usepackage{microtype}

\usepackage{inconsolata}

\usepackage{graphicx}

\usepackage{comment}
\usepackage{booktabs}
\usepackage{fontawesome} 

\usepackage{graphicx}

\title{The AI Language Proficiency Monitor – Tracking the Progress of LLMs on Multilingual Benchmarks}

\author{David Pomerenke\textsuperscript{\normalfont1} \quad Jonas Nothnagel\textsuperscript{\normalfont2} \quad Simon Ostermann\textsuperscript{\normalfont3}\\
    \textsuperscript{1}Bundesministerium für wirtschaftliche Zusammenarbeit und Entwicklung (BMZ)\\ 
    \textsuperscript{2}Gesellschaft für Internationale Zusammenarbeit (GIZ)\\
    \textsuperscript{3}Deutsches Forschungszentrum für Künstliche Intelligenz (DFKI)\\
    {\small \texttt{ david.pomerenke@bmz.bund.de jonas.nothnagel@giz.de simon.ostermann@dfki.de}}
    }

\begin{document}
\maketitle
\begin{abstract}

To ensure equitable access to the benefits of large language models (LLMs), it is essential to evaluate their capabilities across the world’s languages. We introduce the \textbf{AI Language Proficiency Monitor}, a comprehensive multilingual benchmark that systematically assesses LLM performance across up to 200 languages, with a particular focus on low-resource languages. Our benchmark aggregates diverse tasks including translation, question answering, math, and reasoning, using datasets such as FLORES+, MMLU, GSM8K, TruthfulQA, and ARC. We provide an open-source, auto-updating leaderboard and dashboard that supports researchers, developers, and policymakers in identifying strengths and gaps in model performance. In addition to ranking models, the platform offers descriptive insights such as a global proficiency map and trends over time. By complementing and extending prior multilingual benchmarks, our work aims to foster transparency, inclusivity, and progress in multilingual AI. The system is available at \url{https://huggingface.co/spaces/fair-forward/evals-for-every-language}.

\end{abstract}

\section{Introduction}

To make the benefits of large language models (LLMs) available to everyone, LLMs must speak and understand everyone’s language. While a growing range of multilingual LLMs exist, mainstream LLM development still predominantly focuses on English. In contrast to this, a majority – close to 80\% – of the world’s population does not speak English \cite{ethnologue2025}. In response to this disparity, a significant number of multilingual resources for evaluation and training have been developed for many languages, both for pretraining \cite{mayer-cysouw-2014-creating,conneau-etal-2020-unsupervised,abadji-etal-2022-towards,weber2024redpajama} and finetuning LLMs \cite{fu-etal-2022-polyglot,chiang2023vicuna,asai-etal-2024-buffet}. With this work, we present the \textit{AI Language Proficiency Monitor}, an attempt to compile a number of such multilingual evaluation datasets into a comprehensive benchmark for evaluating the performance of current AI models on every language that has available benchmarking data. \textbf{We specifically focus on low-resource languages.}

\begin{figure}[t]
    \centering
    \includegraphics[width=\linewidth]{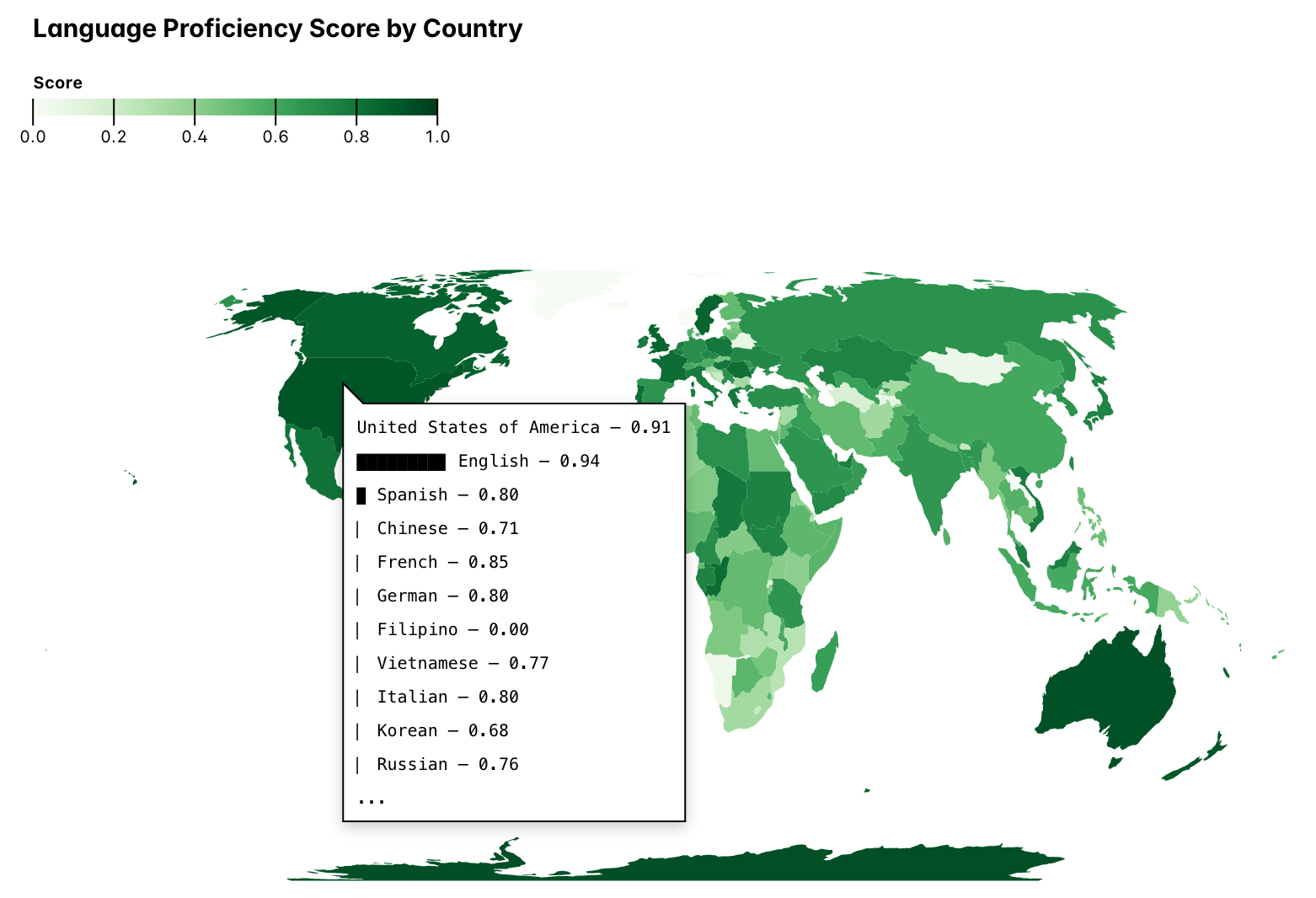}
    \caption{The map view of the language proficiency monitor.}
    \label{fig:map}
\end{figure}

The benchmark is accompanied by an open-source dashboard and is designed to support three primary use cases:
\begin{itemize}
    \item Researchers and model developers can benchmark their systems using the ``AI Language Proficiency Score,'' a multi-task multilingual performance metric;
    \item Practitioners can identify the most suitable models for specific languages and tasks;
    \item Policymakers and donor agencies can identify languages where capacity-building is most urgent.
\end{itemize}
With this design, we aim to provide impact beyond academia by supporting both private- and public-sector decision-making, with a particular emphasis on underserved communities.

For benchmarking we rely on \textbf{FLORES+} (\citet{goyal-etal-2022-flores}, translation, classification) and multilingual versions of \textbf{MMLU} (\citet{hendrycks2020measuring}, multi-domain question answering), \textbf{ARC} (\citet{clark2018think}, advanced question answering), \textbf{GSM8K} (\citet{cobbe2021training}, math problems), and \textbf{TruthfulQA} (\citet{lin-etal-2022-truthfulqa}, factuality). Depending on the task, 100-200 languages are evaluated; these cover 80-95\% of the world’s speakers, and include many low-resource languages. The monitor evaluates state-of-the-art models from commercial AI providers, as well as relevant open-weights models. The presented pipeline auto-evaluates daily, always including the latest relevant models.

In addition to an interactive scoreboard, we present descriptive analyses: A global map of country-level scores of AI language proficiency; progress over time for high-resource vs. low-resource languages; and the impact of a language’s number of speakers, GDP, and data availability on AI language proficiency. Our goal is to also address non-academic users, who often struggle to use and interpret academic benchmarks \cite{ethayarajh2020utility}.

In comparison to existing research and systems, we make the following contributions: While comprehensive multilingual benchmarks such as FLORES+ have already been used for evaluating models at singular points in time, our system continually applies them to the evolving landscape of LLMs. This is similar to the OpenLLM leaderboard\footnote{\url{https://huggingface.co/spaces/open-llm-leaderboard/open_llm_leaderboard}} and the underlying EleutherAI LM Evaluation Harness \cite{eval-harness}, but with a focus on systematic multilingual evaluation. Regional leaderboards for African and European languages pursue the same focus (see \autoref{sec:bg}) and also leverage machine translation to achieve comprehensiveness across the languages they cover; our system takes inspiration from them, while scaling to a global level, including more and up-to-date models, and providing a user-friendly interface with language-specific leaderboards and visuals.

While our analysis already covers the most popular models and the largest languages, it is far from comprehensive. We thus hope that our initiative sparks the interest of regional AI developers to submit their own finetuned AI models, and to integrate datasets for further tasks and languages.

\section{Background and Related Work}\label{sec:bg}

\subsection{Multilingual Evaluation Benchmarks}
A number of multilingual LLM benchmark exists. Notably, all of these benchmarks either come with a very specific regional focus, or a low coverage of low-resource languages. 
A first example is the \textit{European LLM Leaderboard}\footnote{\url{https://huggingface.co/spaces/Eurolingua/european-llm-leaderboard}} \cite{thellmann2024multilingualllmevaluationeuropean}. It is based on the OpenGPT-X LM-evaluation-harness\footnote{\url{https://github.com/OpenGPTX/lm-evaluation-harness}} and include NLU and machine translation evaluations in 21 European languages (Irish, Maltese, Croatian are missing); on publicly available data sets such as MMLU \cite{hendrycks2020measuring}, HellaSwag \cite{zellers2019hellaswag}, and GSM8K\cite{cobbe2021trainingverifierssolvemath}. Notably, only European languages are covered. 

Another example is the \textit{Multilingual LLM Evaluation}\footnote{\url{https://huggingface.co/collections/CohereLabs/multilingual-llm-evaluation-674ebf8e49e5af2efe021c8d}} benchmark collection by Cohere.AI. The benchmark includes a collection of datasets like \textit{Global-MMLU} or \textit{aya-redteaming}. While being an exhaustive collection of multilingual evaluation datasets, the benchmark doesn't provide a unified evaluation interface or visualizations. 

\textit{MEGA} (Multilingual Evaluation of Generative {AI}) \cite{ahuja-etal-2023-mega} is a collection of standard NLP benchmarks, covering 16 NLP datasets across 70 typologically diverse languages. \textit{MEGAVERSE} \cite{ahuja-etal-2024-megaverse} extends this to 22 datasets covering 83 languages, as well as some multimodal benchmarks. Both benchmarks are most similar to ours in terms of coverage, but still lack behind a more holistic evaluation of languages (with our benchmark comprising more than twice the amount of languages), and they come without an explicit leader board or visualizations, restricting their usefulness for non-technical practitioners.

A prominent example for a more local and regional benchmark is \textit{AfroBench} \cite{ojo2025afrobenchgoodlargelanguage}.\footnote{\url{https://mcgill-nlp.github.io/AfroBench/leaderboard.html}}
the benchmark encompasses data from 64 African languages for nine natural language understanding datasets, six text generation datasets, six knowledge and question answering tasks, and one mathematical reasoning task. Naturally, it focuses on African languages only.

\subsection{Multilingual and non-English Language Models}
Multilingual large language models (MLLMs) are able to process and generate text across multiple languages, enabling cross-lingual applications such as translation, retrieval, and reasoning. Early examples of multilingual language models include mBERT \cite{devlin-etal-2019-bert} and XLM-R \cite{conneau-etal-2020-unsupervised}, encoder-based models that are small by modern standards. There also exist some embedding-based works, especially for low-resource languages \cite{gremlin}. For LLMs, the most famous examples include the LLama 3 model family \cite{grattafiori2024llama} and Gemma 3 \cite{team2025gemma}, two types of open-weights models. 

\citet{qin2025survey} introduce a comprehensive taxonomy of MLLMs based different techniques to bridge language gaps. These include (1) \textit{shared vocabulary/token alignment}, where subword tokenizers like SentencePiece are used across languages, which sometimes results in tokenization biases that disadvantage low-resource languages \cite{rust2021good}; (2) \textit{architecture-level alignment}, where model parameters are shared across languages and sometimes augmented with language-specific adapters to balance efficiency and specialization \cite{pfeiffer2020adapterfusion}; (3) \textit{representation-level alignment}, where multilingual embeddings are directly aligned using training objectives such as contrastive or parallel-data losses \cite{artetxe2018massively}; and (4) \textit{post-hoc alignment}, in which independently trained monolingual models are projected into a shared embedding space, showing emergent alignment even without explicit multilingual training \cite{conneau-etal-2020-emerging}. Each method involves trade-offs between scalability, fairness, and computational cost. 
\section{System Design and Implementation}

\subsection{Dataset Collection and Translation}

We collect multilingual benchmark datasets for common NLP tasks comprising translation, text classification, question answering, and maths. We focus on datasets with parallel data for as many languages as possible, so that meaningful comparisons of results across languages are possible. The selected datasets are listed in \autoref{tab:benchmarks}. The table also gives information about how the translation was done (by humans vs. machine-translated). 

\begin{table*}[ht]
\small
\centering
\label{tab:benchmarks}
\begin{tabular}{lcrc}
\toprule
                                  & \textbf{\#Languages} & \textbf{Translation} & \textbf{Authors} \\
\midrule
\multicolumn{4}{l}{\textbf{Translation}} \\
FLORES+                           & 200            & \faUser       & \citet{costa-jussaScalingNeuralMachine2024a} \\
\midrule
\multicolumn{4}{l}{\textbf{Classification}} \\
SIB-200                           & 200            & \faUser       & \citet{adelaniSIB200SimpleInclusive2024}     \\
\midrule
\multicolumn{4}{l}{\textbf{Q\&A: Multitask Language Understanding}} \\
MMLU (Original)                   & 1              & --            & \citet{hendrycksMeasuringMassiveMultitask2021} \\
\addlinespace
MMMLU                             & 14             & \faUser       & \href{https://huggingface.co/datasets/openai/MMMLU}{OpenAI} \\
GlobalMMLU                        & 42             & \faUser\faCog        & \citet{singhGlobalMMLUUnderstanding2025}     \\
AfriMMLU                          & 17             & \faUser       & \citet{adelaniIrokoBenchNewBenchmark2025}     \\
OkapiMMLU                         & 26             & \faCog        & \citet{laiOkapiInstructiontunedLarge2023}     \\
MMLU Auto-Translated              & 61            & \faCog        & --                                           \\
\midrule
\multicolumn{4}{l}{\textbf{Q\&A: ARC Challenge}} \\
ARC (Original)                    & 1              & --            & \citet{clarkThinkYouHave2018}                 \\
\addlinespace
Uhura ARC Easy                    & 6              & \faUser       & \cite{bayesUhuraBenchmarkEvaluating2024}      \\
Okapi ARC Challenge               & 31             & \faCog        & \citet{laiOkapiInstructiontunedLarge2023}     \\
ARC-X                             & 20             & \faCog        & \citet{thellmannMultilingualLLMEvaluation2024} \\
ARC Auto-Translated               & 94            & \faCog        & --                                           \\
\midrule
\multicolumn{4}{l}{\textbf{Truthfulness}} \\
TruthfulQA                        & 1              & --            & \citet{lin-etal-2022-truthfulqa} \\
\addlinespace
Uhura TruthfulQA                  & 6              & \faUser       & \cite{bayesUhuraBenchmarkEvaluating2024} \\
Okapi TruthfulQA                  & 31             & \faCog        & \citet{laiOkapiInstructiontunedLarge2023} \\
TruthfulQA-X                      & 94             & \faCog        & \citet{thellmannMultilingualLLMEvaluation2024} \\
\midrule
\multicolumn{4}{l}{\textbf{Grade School Math}} \\
GSM8K (Original)                   & 1              & --            & \citet{cobbeTrainingVerifiersSolve2021}      \\
\addlinespace
MGSM                              & 10             & --            & \citet{shiLanguageModelsAre2022}             \\
AfriMGSM                          & 18             & \faUser       & \citet{adelaniIrokoBenchNewBenchmark2025}     \\
GSM8K-X                           & 20             & \faCog        & \citet{thellmannMultilingualLLMEvaluation2024} \\
MGSM Auto-Translated              & 73            & \faCog        & --                                           \\
\bottomrule
\end{tabular}
\caption{Benchmark datasets by task. The table cover the number of covered languages and information on whether the translations between languages were done by humans (\faUser) or automatically (\faCog).}
\label{tab:benchmarks}
\end{table*}

The most comprehensive of the integrated datasets is FLORES+ with 200 languages, along with the derived SIB-200 dataset, where the same sentences are classified into topic categories. For question answering and mathematics, benchmark datasets that were originally published in English have been translated into additional languages, either by humans or by translation software, as specified in \autoref{tab:benchmarks}. When multiple versions of the data sets cover the same language, we prefer those that are human-translated. When a dataset is present in multiple scripts for the same language, we retrieve script user population data for the relevant language from \textit{Unicode CLDR}\footnote{\url{https://cldr.unicode.org/}} and only incorporate the version with the more common script (the less common script is typically a transliteration from the original script). The Masakhane datasets (AfriMMLU, AfriMGSM, Uhura ARC Easy) only cover a subset of the respective original English datasets; therefore, we restrict all corresponding datasets to exactly the same rows as the Masakhane datasets to ensure that our combined dataset is parallel. 

We retrieve the 100 most spoken languages in the world from CLDR and complement all languages that are not covered by the existing benchmark datasets by using machine translation. While machine translation is inferior to human translation for the purpose of benchmarking, it is a viable alternative for languages without benchmark data, and has been used before by GlobalMMLU \cite{singhGlobalMMLUUnderstanding2025}, Okapi \cite{laiOkapiInstructiontunedLarge2023}, and EU-20 \cite{thellmannMultilingualLLMEvaluation2024}. We select \textit{Google Neural Machine Translation} (Google Cloud Translate v2), which performs better than any general-purpose LLM in our FLORES+ translation evaluation, and use it to translate benchmark questions and answers into all missing languages.


\subsection{Task Implementation and Evaluation}

All model evaluations in the leaderboard are conducted in a few-shot manner. Translated prompt instructions are not provided by the selected datasets, and human translations at such large scale are out of scope for our project. Machine-translating the prompt into different languages would have been an alternative, but potential translation errors would have an overproporionate impact on performance. Thus, we instead use minimal language-agnostic prompts along with few-shot prompting, to demonstrate rather than explain the tasks to the prompted models. For the mathematics task, we specify a response format "<reasoning> \#\#\#\# <number>" in the prompt to give all models a reasoning scratchpad. Examples of few-shot prompts are provided on Github.\footnote{\url{https://github.com/datenlabor-bmz/evals-for-every-language/blob/main/notes/prompt-examples.md}}

All classification and question answering tasks are given in a multiple choice format with one correct answer, and they are evaluated using accuracy. The math tasks are evaluated using the accuracy of the final number. For translation, evaluation is less straightforward: The often-used BLEU metric \cite{papineni-etal-2002-bleu} may be biased towards giving worse scores to languages with coarser tokenization; BertScore \cite{zhang2019bertscore} and its multilingual versions may be biased towards giving a higher score towards languages seen during training. We therefore use the SpBLEU metric \cite{goyal-etal-2022-flores}, which relies on a SentencePiece classifier for tokenization, which has been equally trained on all languages contained in the FLORES+ dataset. Moreover, we distinguish between the task of translating \textit{from} a language of interest to a set of other languages (which is sampled representative of speaker populations) and the task of translating from a representative set of languages \textit{to} a language of interest. While the latter is potentially sensitive to language-specific biases of the evaluation metric, the former task (translating \textit{from} a given language) is more comparable, since the output sentences on which the evaluation metric operates are stable across evaluated languages.

\subsection{Models and Inference}

To make the leaderboard as useful as possible, it should always include up-to-date models, focus on those that are also most relevant to the community of practitioners, and include open-weights as well as API-only models. We use \href{https://openrouter.ai}{OpenRouter} both for retrieving information on historically and currently popular models, and for performing inference. Our pipeline updates daily via Github Actions to retrieve the latest models and run evaluations on them automatically. OpenRouter has a central functionality to only use endpoints where providers are not allowed to train on user data, which is crucial for avoiding data contamination when performing benchmarking via APIs.

For non-commercial, open-source models - often of particular relevance to local communities due to their task- or language-specific design and lower operational costs - we rely on HuggingFace’s inference API. This approach also provides a simple and inclusive mechanism for users to submit and benchmark their own models, as anyone can publish endpoints on HuggingFace. At present, we ask users to submit a request via an online form and share their model endpoint to be included in the automated evaluation runs. Future versions of the leaderboard will support a fully automated submission and integration process.

\subsection{The Language Proficiency Score}

For our leaderboard we normalize the scores of all tasks using min-max-normalization, so that scores are more comparable across tasks.

We define an aggregate metric, the \textbf{Language Proficiency Score}, as the mean of the individual task metrics. This metric provides an overall rating of an AI models understanding of a given language, and is intended for reporting on multilingual abilities of new AI models. On our leader board, we report aggregate Language Proficiency Scores both for each AI model (averaging over all languages) and for each language (averaging over all models).
\begin{figure}
    \centering
    \includegraphics[width=\linewidth]{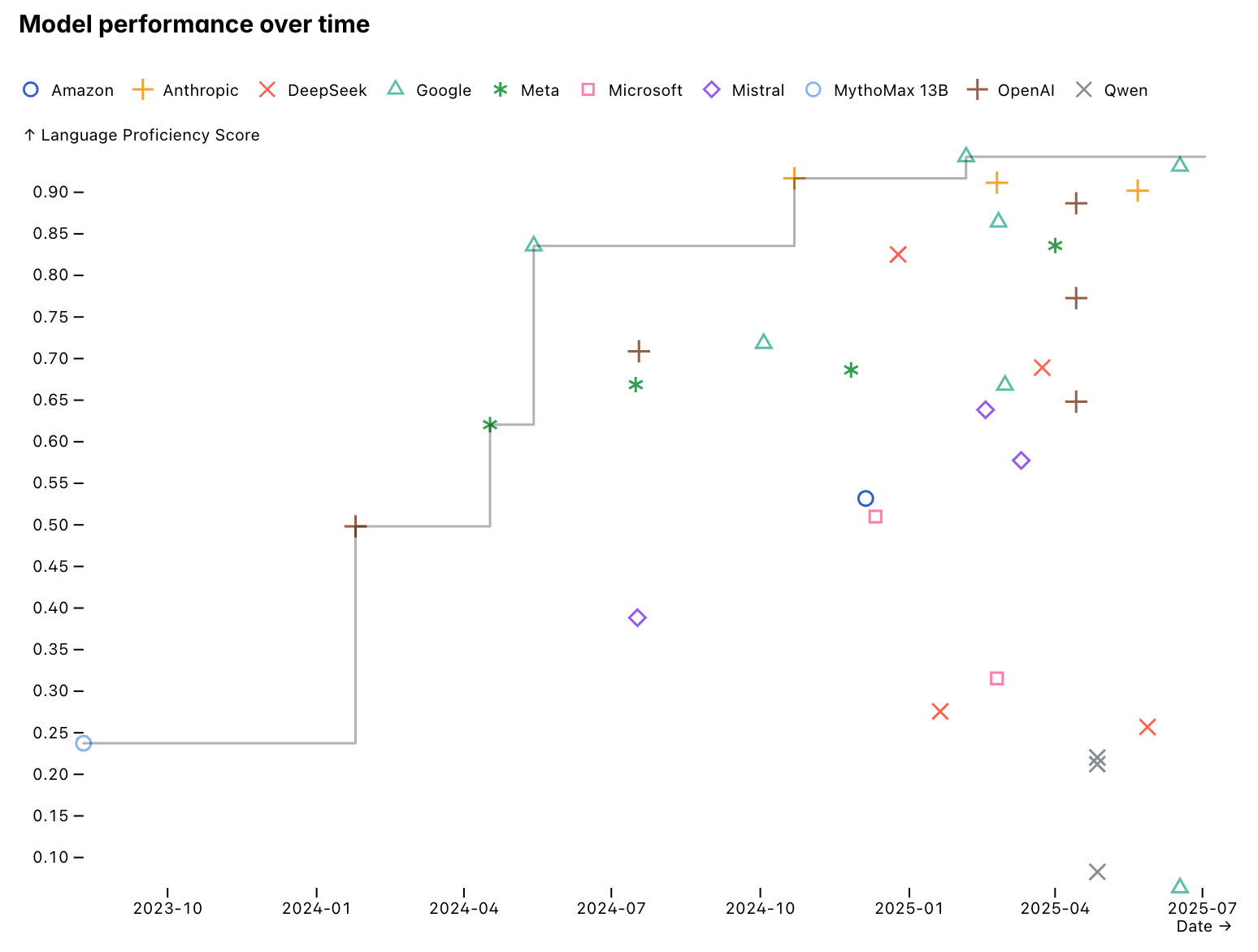}
    \caption{Language proficiency scores by different models over time. The line shows the highest achieved score by any model at a given date.}
    \label{fig:scores-over-time}
\end{figure}

\begin{figure}
    \centering
    \includegraphics[width=\linewidth]{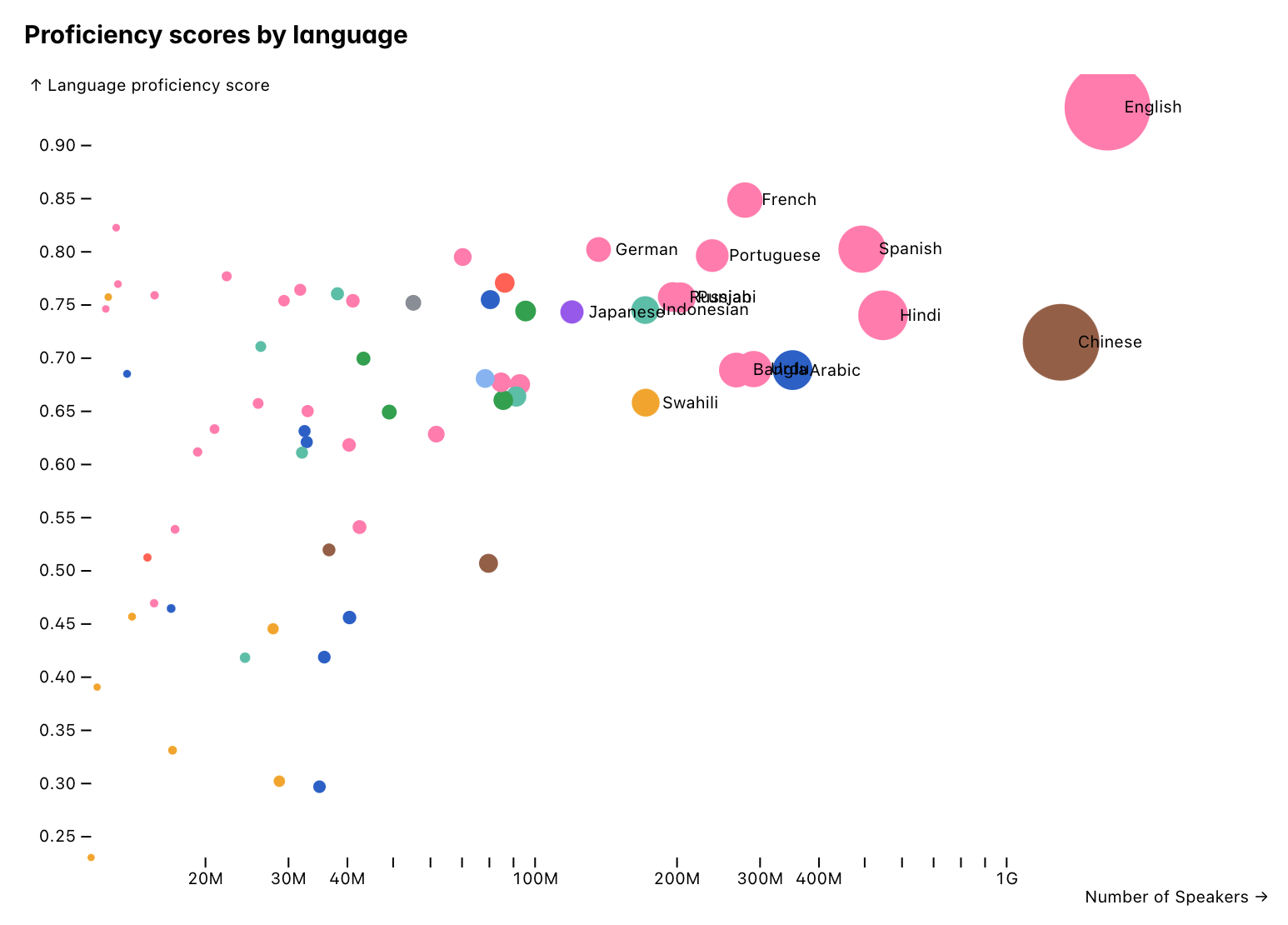}
    \caption{Language Proficiency Score by language.}
    \label{fig:scores-by-language}
\end{figure}

\subsection{Web Application}

Our web application is hosted on Huggingface, where it is linked to existing models and datasets. The web app consists of four parts:

\begin{itemize}
    \item \textbf{Model leaderboard.} This table ranks LLMs by their language proficiency and provides the Language Proficiency Score as well as scores for each task. The table can be filtered by open-weights vs API-only models, by cost per output tokens, and for the open models also by parameter size; this metadata is retrieved from OpenRouter and Huggingface. By default, the leaderboard shows scores that are aggregated across all evaluated languages. The leaderboard picker allows users to switch instead to leaderboards for a specific language of interest.
    \item \textbf{Language scoreboard.} This table shows information on languages and how proficient AI models are in them. Languages are sorted by overall number of speakers, and may be filtered by language family. The language Proficiency Score and task-specific scores are displayed. The user may select one or multiple languages, and the model leaderboard will focus on the selected languages.
    \item \textbf{Dataset infoboard.} This table presents data on the benchmark datasets that the system uses for evaluation, and they can be filtered by task, number of languages, and human- vs machine-translated datasets.
    \item \textbf{Visuals board.} This view provides interactive plots, including a map, trends over time, and a cost-effectiveness comparison (see \autoref{fig:map}, \autoref{fig:scores-by-language}, \autoref{fig:scores-over-time}).
\end{itemize}

A screenshot of the system is included in \autoref{sec:appendix}.
Code and evaluation pipeline are released under the MIT License.\footnote{\url{https://opensource.org/licenses/MIT}. Copyright © 2025 Datenlabor BMZ, GIZ, and DFKI.} All code and documentation are available at \url{https://github.com/datenlabor-bmz/evals-for-every-language}.

\section{Evaluation}

\subsection{Model and Language Performance}
\autoref{fig:scores-by-language} shows the language proficiency scores by language. The size of the dots and the positioning on the x axis indicates the number of speakers of the respective language. The graph confirms already known trends: English as the language with most resources employs the highest language proficiency scores. Large European Languages like French, Spanish, Portuguese, or German show high scores in spite of comparably fewer speakers. Larger low to mid resource languages like Swahili show comparably bad scores in spite of many speakers. \autoref{fig:scores-over-time} shows language proficiency scores by different models over time, showcasing that Google models currently perform best.

\subsection{Qualitative Stakeholder Feedback}
In order to get an idea about the usefulness of our tool for the relevant stakeholder, we conducted a small qualitative survey comparing our benchmark along other freely available benchmarks (AfroBench, European LLM Leaderboard, Multilingual MMLU Benchmark Leaderboard, Multilingual Reasoning Leaderboard) with one stakeholder each from Industry, SME, Public Sector, and NGO. In comparison, the stakeholders liked the simplicity of the AI monitor use and the focus on non-European languages, while they criticized that still only academic benchmarks were used, the performance on which doesn't necessarily correlate with downstream application performance in a language. While they liked the language coverage and ease of use of the dashboard, they still lacked cross-lingual trends and would have expected an even larger coverage of tasks - which we will account for in future versions of the dashboard.

\section{Conclusion}

\subsection{Limitations}
A key limitation of our work arises from the inherent challenges of using  benchmarks to evaluate large language models. As widely discussed in recent literature, for example \cite{raji2021ai} \cite{asai-etal-2024-buffet}, benchmarks often offer only a very specific snapshot of capabilities and may not reflect real-world utility, particularly across diverse, low-resource contexts and use-cases. Our monitor naturally inherits these limitations, as it consolidates and standardizes many existing benchmarks into a unified leaderboard.

In addition, both resource constraints and the prototype nature of our system currently limit the scope of our benchmarking runs. Evaluating all models across the full benchmark—spanning up to 200 languages and multiple tasks—would require substantial compute. At this stage, our goal is to demonstrate the feasibility of the approach before exhaustively benchmarking every model. As a result, our automated runs currently sample 10 instances per combination of model, task, and language. We plan to scale up coverage in future iterations through additional compute resources and community contributions. 

\subsection{Future Directions}

While our monitor aims to offer a broad, inclusive view of multilingual LLM performance, particularly for low-resource languages, its current coverage can still be extended. More localized models, benchmarks, and underrepresented languages could still be added to provide a more complete and impactful overview. To address this, we plan to leverage our open-source framework to iteratively include these together with regional communities and researchers.

\subsection{Summary}
We introduced the AI Language Proficiency Monitor, a comprehensive, continually updated benchmark that tracks the multilingual capabilities of large language models across up to 200 languages. By combining diverse evaluation datasets and presenting results through an accessible and open-source dashboard, our system enables developers, researchers, and policymakers to identify performance disparities and support equitable language representation. Our approach provides a scalable foundation for inclusive, transparent, and actionable multilingual AI evaluation. We invite the broader community to contribute new models, tasks, and languages to further strengthen this shared resource.

\bibliography{custom}

\clearpage

\appendix

\onecolumn
\section{Appendix}
\label{sec:appendix}

\includegraphics[width=\linewidth]{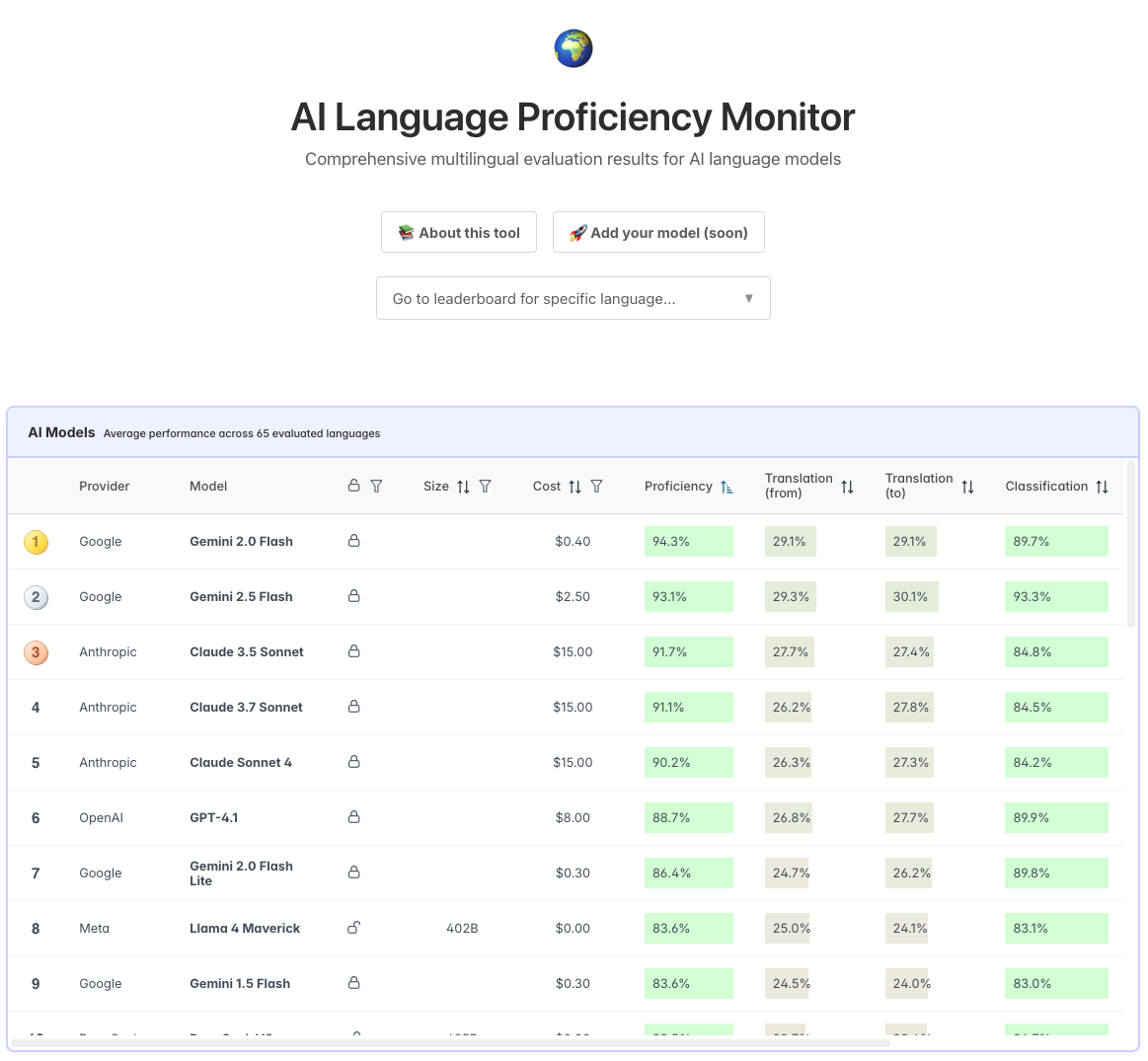}
Screenshot of the model leaderboard, showing aggregate score across languages per model, overall and by task. The user can filter for specific types of models, or switch to a language-specific leaderboard for any of the evaluated 200 languages.

\end{document}